\documentclass[
]{ceurart}

\usepackage{soul}
\usepackage{comment}

\usepackage{booktabs}
\usepackage{scalerel,graphicx,xparse}
\NewDocumentCommand\emojismile{}{{\includegraphics[scale=.04]{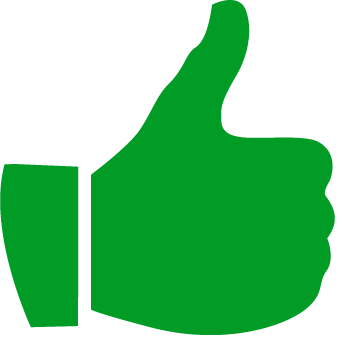}}}
\NewDocumentCommand\emojifrown{}{{\vspace{1pt}\includegraphics[scale=.04]{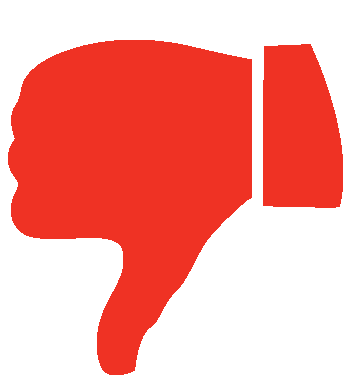}}}

\begin{document}

\copyrightyear{2021}
\copyrightclause{Copyright for this paper by its authors.
  Use permitted under Creative Commons License Attribution 4.0
  International (CC BY 4.0).}

\conference{Deep Learning for Knowledge Graphs (DL4KG)}


\title{Language Models As or For Knowledge Bases}

\author[1]{Simon Razniewski}[%
email=srazniew@mpi-inf.mpg.de,
]
\address[1]{Max Planck Institute for Informatics}

\author[1,2]{Andrew Yates}[%
email=ayates@mpi-inf.mpg.de,
]
\address[2]{University of Amsterdam}

\author[3]{Nora Kassner}[%
email=kassner@cis.lmu.de,
]
\address[2]{LMU Munich}

\author[1]{Gerhard Weikum}[%
email=weikum@mpi-inf.mpg.de,
]

\newcommand{\GW}[1]{{\color{blue}{GW: #1}} }
\newcommand{\sr}[1]{{\color{green}{SR: #1}} }

\maketitle


\begin{abstract}
Pre-trained language models (LMs) have recently gained attention for their potential as an alternative to (or proxy for) explicit knowledge bases (KBs).
In this position paper, we examine this hypothesis, identify strengths and limitations
of both LMs and KBs, and discuss the complementary nature of the two paradigms.
In particular, we offer qualitative arguments that latent LMs are not suitable {\em as} a substitute for explicit KBs, but could play a major role 
{\em for} augmenting and curating KBs.
\end{abstract}

\setcounter{page}{1}

\vspace*{-0.5cm}
\section{Introduction}

The ability of pre-trained contextual language models (LMs) to capture and retrieve factual knowledge has recently stirred discussion as to what extent LMs could be an alternative to, or at least a proxy for, explicit knowledge bases (KBs).
LMs, such as BERT~\cite{bert}, GPT~\cite{gpt}
or T5~\cite{raffel2019exploring} are huge transformer-based neural networks trained in a self-supervised manner on huge text corpora, in order to predict sentence completions or masked-out text parts.
In a setting called (masked) prompting or probing
\cite{liu2021pre}, these LMs complete a text sequence intended to elicit a relational assertion for a given subject. For example, GPT-3 correctly completes the phrase 
\textit{``Alan Turing was born in''} with 
\textit{``London''}, which can be seen as yielding a subject-predicate-object triple 
{\small\tt $\langle$ Alan Turing, born in, London $\rangle$}.
%

Starting from the 
LAMA probe \cite{petroni2019language}, many works have explored whether this LM-as-KB paradigm could provide an alternative to structured knowledge bases
such as Wikidata. 
Exemplary analyses investigated the inclusion of entity information \cite{poerner2020ebert}, how to turn LMs into structured KBs \cite{wang2021language}, and how to incrementally add knowledge without side effects \cite{wang2021kadapter}.
Other work studied how accuracy relates to the neural network's storage 
capacity \cite{heinzerling-inui-2021-language} and whether
QA performance scales with model size \cite{roberts2020knowledge}.
 Another focus area is how LMs-as-KBs can be further augmented with a text retrieval component, to include informative passages (e.g., from Wikipedia) \cite{petroni2020context,guu2020realm,lewis2021retrievalaugmented}.

Although most works 
make their speculative nature clear
(e.g., the title of \cite{petroni2019language} ends with
a question mark),
there is an implicit suggestion that LMs could replace structured KBs.
On the other hand, NLP-centric works
 have identified various kinds of inconsistencies in LM outputs \cite{elazar2021measuring} or questioned their quantitative performance \cite{cao2021knowledgeable}.



This 
paper discusses 
the potential of LMs {\em as} KBs
and its ``softer'' variation of
LMs {\em for} KBs.







\section{Background}

LM-as-KB refers to efforts to use an LM as a source of world knowledge, as proposed by \cite{petroni2019language}. 
The knowledge representation is inherently latent, given by the entirety of the neural network's parameter values (in the billions).
LMs in general have greatly advanced tasks like text classification, machine translation, information retrieval, and question answering (see, e.g., survey \cite{recenttrends}).

KBs, on the other hand, have been steadily advanced since the mid 2000s (with early works like DBpedia, Freebase and Yago) \cite{razniewski2020structured}. They represent knowledge in the form of subject-predicate-object (SPO) triples along with qualifiers for non-binary statements.
KBs have become key assets in major industry applications 
\cite{noy2019industry,weikum2021machine}, including search engines.
A major issue for ongoing KB research is
quality assurance as the KB is grown and maintained.
This includes human-in-the-loop approaches 
throughout the KB life-cycle
\cite{jamietaylor,piscopo2019we,shenoy2021study}.
%

All LM-as-KB examples that follow are based on the GPT-3 daVinci model~\cite{gpt}, one of the largest pre-trained LMs as of October 2021. 

\begin{table}[t]
\scalebox{0.85}{
\begin{tabular}{@{}lll@{}}
\toprule
 & \textbf{LM-as-KB} & \textbf{Structured KB} \\ \midrule
\textbf{Construction} & Self/Unsupervised 
\emojismile & \begin{tabular}[c]{@{}l@{}}Manual or semi-automatic
\emojifrown 
\end{tabular} 
\\
\textbf{Schema} & Open-ended 
\emojismile & Typically fixed \emojifrown \\
\textbf{Maintenance} & & \\[-4pt]
  \ \ - adding facts & Difficult, unpredictable side effects \emojifrown & Easy \emojismile\\[-4pt]
  \ \ - correcting/deleting & Difficult \emojifrown & Easy \emojismile  \\
\textbf{Knows what it knows} & No, assigns probability to everything \emojifrown & Yes, content enumerable \emojismile \\
\textbf{Entity disambiguation} & No/limited \emojifrown & Common \emojismile \\
\textbf{Provenance} & No \emojifrown & Common \emojismile\\
\bottomrule
\end{tabular}
}
\label{tbl:comparison}
\caption{Differences of LMs-as-KBs and structured KBs}.\\
\end{table}


\section{LM-as-KB}

%


\subsection{Intrinsic Considerations}

The following are principal differences between LMs-as-KBs and structured KBs.


\vspace{0.1cm}
\noindent         \textbf{Predictions vs. lookups}: 
    While content of structured KBs can be explicitly looked up, LMs have a latent representation and output probabilities at probing time. 
    This has the advantage of not requiring any schema design upfront. However, it implies that it is not possible to enumerate the knowledge stored in an LM, nor can we look up whether a certain fact is contained or not.
    %
    %
    For predictions with very high confidence scores, this is still viable. However, even top-ranked predictions often have low scores and near-ties. Properly calibrating scores and thresholding is a black art.
    
    \textit{Example: GPT-3 does 
    not have tangible knowledge that Alan Turing was born in London; it merely assigns this a high confidence of 83\%.
    Yann LeCun, on the other hand, is given medium confidence in being a citizen of France and Canada (67\% and 26\%), but he actually has French and USA citizenship, not Canadian. The LM assigns USA a very low score. The Wikidata KB, on the other hand, only states his French citizenship, not USA. Wikidata is incomplete, but it does not contain any errors.
    %
    }
    \vspace{0.1cm}


\noindent         \textbf{Statistical correlations vs. explicit knowledge}: Errors made by LMs-as-KBs are not random, but exhibit systematic biases \cite{parrots,cao2021knowledgeable} 
    due to frequent values and co-occurrences (including indirect co-occurrences captured latently).
    
    \textit{Example: 
    %
    %
    When prompting GPT-3 for awards won by Alan Turing,
    its top-confidence prediction is the Turing Award, and lower-ranked outputs include ``Nobel Prize'' and ``the war'' (none of them correct).
    }
    \vspace{0.1cm}


\noindent         \textbf{Awareness of limits}: In KBs, 
    absence of facts is explicit and easy to assert. Wikidata even supports a way of stating non-existence (no-value statements) to impose a local-closed-world view while following a general open-world assumption \cite{arnaout2021negative}.
    LM's latent representations inherently lack awareness of cases where no object exist, and so they easily produce non-zero or even high scores for incorrect assertions. 
    
    \textit{Example: 
    Alan Turing was homosexual and never married. When prompting GPT-3 with the phrase ``Alan Turing married'', the top prediction is
    ``Sara Lavington'' with score 21\%,
    and for the prompt ``Alan Turing and his wife'' it is ``Sara Turing'' (his mother's name).
    This is a case of LM hallucination 
    \cite{hallu1,hallu2}.
    In contrast, Wikidata has an explicit statement {\small\tt $\langle$ Alan Turing, spouse, no value $\rangle$}
    denoting that he was unmarried.
    }
    \vspace{0.1cm}  
    

\noindent         \textbf{Coverage}: 
    The scope of KBs is usually limited by the fixed set of predicates specified in the KB schema. These can be hundreds (or even a few thousands) of interesting relations, but will hardly ever be complete. In particular, ``non-standard relations'', such as 
    {\small\tt worked with colleague},
    {\small\tt song is about person} (or {\small\tt event}),
    {\small\tt movie based on person's biography}, are missing in all of the major KBs. 
    LMs, on the other hand, latently tap into the full text of Wikipedia, books, news, and more, and are thus able to capture some of these predicates.
    
   \textit{Example: 
   Creatively prompting GPT-3 can yield impressive nuggets of knowledge: 
   the input phrases ``Turing's colleague'' and ``Turing worked with'' result in outputs like
   John Womersley, Hugh Alexander, Gordon Welchman (all correct). 
   Likewise, the prompt ``The Imitation Game film is about the life of'' is completed with the high-confidence output Alan Turing.
   These anecdotes indicate the great power of LMs to go beyond the current scope and coverage of explicit KBs.
   }
    \vspace{0.1cm} 


\noindent          \textbf{Curatability}:
     In structured KBs, a knowledge curator can correct, add or remove assertions. For LMs, this is an open challenge, as these operations require major (non-monotonic) 
    re-training, or the addition of explicit exceptions, which means reverting to a KB~\cite{zhu2020modifying,decao2021editing}. 
    
       \textit{Example:
    For the prompt ``Alan Turing died in the town of'', GPT-3 returns the top prediction ``Warrington'', which is wrong (he died in Wilmslow). 
    The LM does not provide any hint on how to fix this (e.g., by changing the training corpus or parameters), and a knowledge curator has no way to tackle such errors.
    }
    \vspace{0.1cm}


\noindent       \textbf{Provenance}: 
        LMs have no ability to trace their outputs back to specific source documents (and passages) in the training data. 
        KBs, on the other hand, consider
        reference sources as an indispensable pillar of scrutable
        veracity.
      Provenance is crucial for giving explanations to users, including knowledge engineers who maintain the KB and end-users in downstream applications.
      Also, without provenance, LMs have no way of pinpointing an incorrect prediction's root cause and correcting the underlying corpus
    (e.g., removing misleading documents).
    
    \textit{Example: Reconsider the previous example of predicting ``Warrington'' as Turing's death place. The LM itself does not give any cue where this comes from. A diligent and smart Google user could detect a possible origin, namely, news and other reports about a memorial plaque at 2 Warrington Crescent in Maida Vale, London, which is near Turing's birth place. However, the knowledge engineer cannot be certain that this is indeed the culprit.\\
    Correctly predicted facts need explanations, too. For example, the assertion that Turing was engaged with Joan Clarke may appear puzzling given his homosexuality. Pointing to explicit provenance is crucial evidence. 
     }



\subsection{Pragmatic Considerations}

\noindent    \textbf{Entity disambiguation}: 
Although LMs are lauded for their ability to disambiguate words based on context, this happens latently, and there is no easy way to explicitly build this into probing procedures \cite{heinzerling-inui-2021-language,poerner2020ebert}. Consequently, LMs mix up facts from distinct entities that share surface forms. Although structured KBs cannot perform disambiguation on their own either, they can correctly separate assertions.
    
    \textit{Example: 
    GPT-3 completes ``Turing was a famous'' with ``mathematician'', ``computer'', ``code'' etc., stemming from very different entities (including the Turing Machine).
    }
        \vspace{0.1cm}  
    
\noindent    \textbf{Numbers and singletons}:
LMs are good at latently capturing knowledge about predicates with few possible object values, such as nationality or language-spoken. However, when the object values are
    rarely occurring values or even singletons (i.e., occurring only with a single subject), the latent representation is bound to produce errors, and explicit KB storage is superior.
    The same applies to many cases of numeric values, where the value distribution exhibits high entropy.
    
    \textit{Example: 
    For the input ``The Turing Institute's address in London is'', GPT-3 
    returns ``Dilly's Den'' or ``the street called Dilly's Den'' (possibly derived from the famous Piccadilly Circus; the correct value is British Library, 96 Euston Road, London NW1 2DB).
    Rephrasing the prompt does not lead to success either.
    }
        \vspace{0.1cm}  
    
\noindent    \textbf{Subjects with zero or many objects}: 
    An important case where the brittleness of LM predictions becomes a significant problem is when a subject entity has no object value for a given predicate or has many distinct true values.
    The zero-value case often leads to the pitfall that the LM must predict some value. In the many-values case, we could go deep in the ranking of the LM output, but this would usually result in a wild mix of valid and spurious objects, and there is no guideline for how deep we should go into the ranking.
    
    \textit{Example:
    %
    To obtain a list of Turing Award winners, we could prompt GPT-3 with the phrase 
    ``the Turing Award was won by'' and
    receive various predictions like ``Stuart Shieber'', ``John Hopcroft'' and ``Andrew Yao'' (1 false, 2 correct). 
    There are currently 73 winners, all captured in Wikidata. By probing LMs, we would have to go very deep in the prediction ranking to see all of them, but only in a confusing mix of true and false positives.\\
    As for zero-objects, the prompt for ``the first woman on the moon was''
    returns
    Sally Ride, Eileen Collins and others. These are astronauts, but unfortunately, none of them ever landed on the moon. The ground-truth for this example is empty.
    }
     
    

We summarize the main differences in Table~\ref{tbl:comparison}.

\section{LM-for-KB}

Our view of how to harness the great potential of LMs
is to leverage them {\em for KB curation}: maintaining high quality as the KB grows throughout its life-cycle. 
This is a major pain point in KB practice
\cite{jamietaylor,piscopo2019we,shenoy2021study}. For example, when adding new entities, one needs to ensure that they are not duplicates (with slightly different alias names) of existing entities. Likewise, keeping the type system (aka ontology) clean while gradually extending it and ensuring the correctness of new facts are never-ending challenges.

The envisioned role of LMs is to {\em scrutinize SPO assertions} considered for augmenting the KB.
For example, a new fact such as 
{\small\tt $\langle$ Leonardo da Vinci, has won, Turing Award $\rangle$}
could be ``double-checked'' by prompting the LM as to whether it yields high-confidence predictions for this candidate assertion.
This is akin to the way knowledge graph embeddings \cite{wang2017knowledge} have been considered for KB completion. 
However, the key difference is that KG embeddings draw from the KG itself, and thus do not provide complementary evidence. LMs, on the other hand, bring in a new and largely independent perspective, by tapping into text corpora (including Wikipedia, but also news, books etc.). 
If the LM does not yield sufficiently confident support for the candidate fact, it should be refuted.

The converse direction, using LMs to predict assertions and thus generate candidates for new facts, is conceivable too. However, this needs major research to advance  prediction accuracy. 

\section{Conclusion}

In this paper we 
discussed
the strengths and limitations of LMs {\em as} KBs in comparison to structured KBs.
We believe that LMs cannot broadly replace KBs as explicit repositories of structured knowledge. 
While the probabilistic nature of LM-based predictions is suitable for task-specific end-to-end learning, the inherent uncertainty of outputs does not meet the quality standards of KBs. 
LMs cannot separate facts from correlations, and this entails major impediments 
for KB maintenance.
%
We advocate, on the other hand, that LMs can be valuable assets {\em for} KB curation, by providing a ``second opinion'' on new fact candidates or, in the absence of corroborated evidence, signal that the candidate should be refuted.
Other ways of combining the strengths of latent knowledge (LMs) and structured knowledge (KBs)  could be promising as well, such as ``KB-for-LM'' approaches that allow a LM to look up facts from an external memory (e.g.,  \cite{guu2020realm,fevry2020entities,verga2020facts,kassner2021enriching}) and thus have the potential to combine the strengths of both approaches.





\clearpage\newpage
\bibliography{sample-ceur}

\end{document}